\documentclass[sigconf]{acmart}
\copyrightyear{2020}
\acmYear{2020}
\setcopyright{acmcopyright}
\acmConference[SIGIR '20]{Proceedings of the 43rd International ACM SIGIR Conference on Research and Development in Information Retrieval}{July 25--30, 2020}{Virtual Event, China}
\acmBooktitle{Proceedings of the 43rd International ACM SIGIR Conference on Research and Development in Information Retrieval (SIGIR '20), July 25--30, 2020, Virtual Event, China}
\acmPrice{15.00}
\begin{document}
\fancyhead{}

\title{Label-Consistency based Graph Neural Networks \\ for Semi-supervised Node Classification}

\author{
Bingbing Xu, Junjie Huang, Liang Hou, Huawei Shen$^*$,  Jinhua Gao,  Xueqi Cheng}
\affiliation{
\department{$^1$CAS Key Laboratory of Network Data Science and Technology,}
\institution{Institute of Computing Technology, Chinese Academy of Sciences;}
}
\affiliation{\institution{$^2$School of Computer and Control Engineering, University of Chinese Academy of Sciences;}}
\email{{xubingbing, huangjunjie17s, houliang17z, shenhuawei, gaojinhua, cxq}@ict.ac.cn}

\renewcommand{\shortauthors}{Bingbing Xu and Junjie Huang, et al.}

\begin{abstract}
Graph neural networks (GNNs) achieve remarkable success in graph-based semi-supervised node classification, leveraging the information from neighboring nodes to improve the representation learning of target node. The success of GNNs at node classification depends on the assumption that connected nodes tend to have the same label. However, such an assumption does not always work, limiting the performance of GNNs at node classification. In this paper, we propose label-consistency based graph neural network (LC-GNN), leveraging node pairs unconnected but with the same labels to enlarge the receptive field of nodes in GNNs. Experiments on benchmark datasets demonstrate the proposed LC-GNN outperforms traditional GNNs in graph-based semi-supervised node classification. We further show the superiority of LC-GNN in sparse scenarios with only a handful of labeled nodes.

\end{abstract}


\maketitle

{\fontsize{8pt}{8pt} \selectfont
\textbf{ACM Reference Format:}\\
Bingbing Xu, Junjie Huang, Liang Hou, Huawei Shen,  Jinhua Gao,  Xueqi Cheng. 2020. Label-Consistency based Graph Neural Networks for Semi-supervised Node Classification. In \textit{Proceedings of the 43rd International ACM SIGIR Conference on Research and Development in Information Retrieval (SIGIR '20), July 25--30, 2020, Virtual Event, China.} ACM, New York, NY, USA, 4 pages. https://doi.org/10.1145/3397271.3401308}

\renewcommand{\thefootnote}{\fnsymbol{footnote}}
\footnotetext[1]{Corresponding author.}

\section{Introduction}

Owing to the powerful representation capability of the graph, many real-life scenarios, such as transportation networks, social networks, and citation networks, are located in the form of graphs. In these scenarios, graph-based semi-supervised node classification, i.e., classifying nodes in a graph with few labeled nodes, has attracted much attention due to the wide range of applications, e.g., user tagging in social networks and product recommendation. Recently, graph neural networks (GNNs)~\cite{bruna2014spectral,defferrard2016convolutional,xu2019graph,xu22019graph} achieve great success in semi-supervised node classification.

\begin{figure}[t]
\centering  
\includegraphics[width=60mm, height = 50mm]{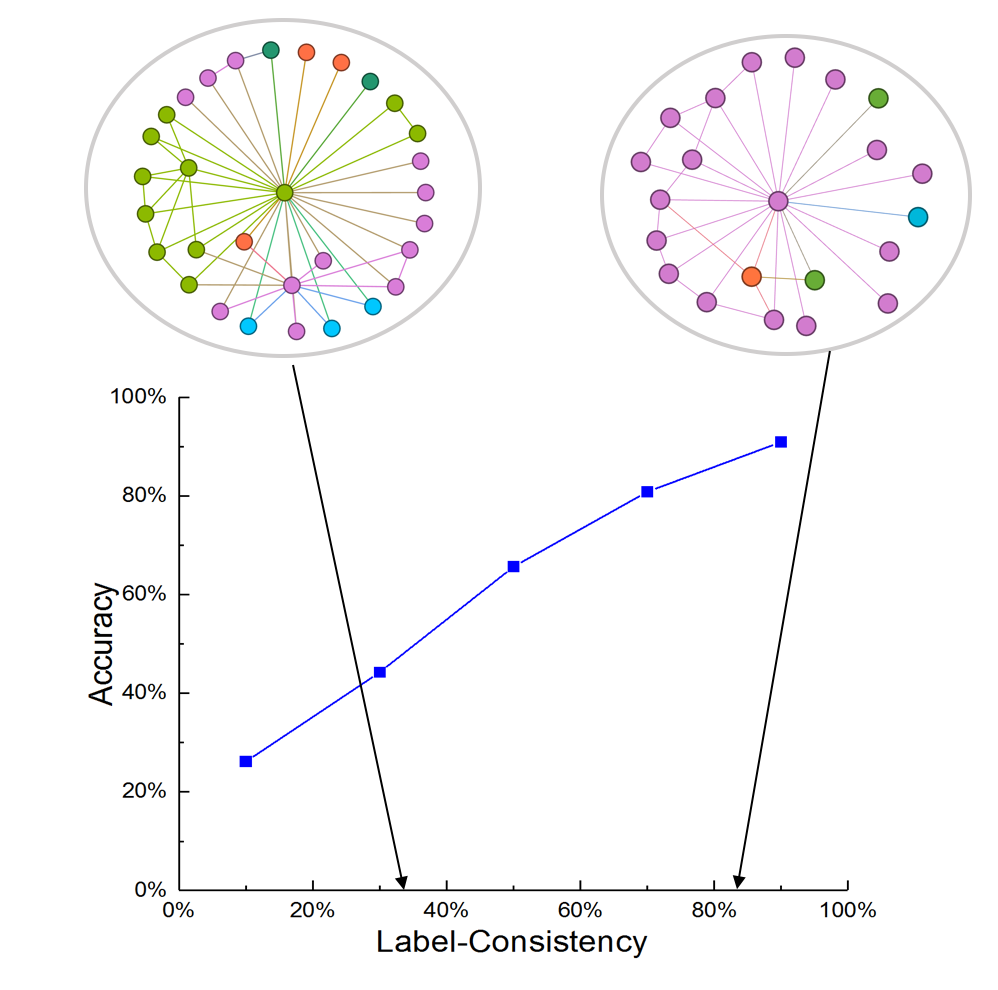}
\caption{Performance of GCN on Cora with different label-consistency: the classification accuracy increases with respect to the label consistency.} 
\label{fig:realideal}
\end{figure}

The success of graph neural networks in node classification depends on the assumption that connected nodes tend to have the same label. With this assumption, graph neural networks leverage the information from neighboring nodes to improve the representation learning of target node. For example, graph convolution network (GCN~\cite{kipf2017semi}) iteratively update each node's representation via aggregating the representation of its neighboring nodes. Symmetric normalized adjacency matrix is used as \emph{aggregation matrix}, which characterizes the importance of neighboring nodes to target node during the process of information aggregation. For GCN, aggregation matrix is solely determined by graph structure, and this limits its capability at aggregating the information from nodes with similar features or attributes. To combat the shortcoming of GCN, graph attention network (GAT~\cite{velickovic2017graph}) is proposed. GAT defines a novel aggregation matrix via a self-attention mechanism, quantifying the importance of neighboring node by its similarity to target node in terms of its feature vector or representation, and adjacency matrix is used as a mask. In this way, GAT leverages both graph structure and node features to define aggregation matrix. However, for both GCN and GAT, only neighboring nodes are considered when defining aggregation matrix. There existed some methods incorporating label consistency and feature correlations~\cite{jin2019graph, qu2019gmnn}. Besides, previous proposed methods also attempt to combine label propagation with neighbor aggregation to achieve the goal~\cite{jiang2019semi, prasad2019glocal}. However, these methods relied on the original features and labels, resulting in noise and inefficiency. Consequently, these traditional graph neural networks cannot take advantage of unconnected nodes but with the same label to improve the representation learning of target node. Furthermore, the aggregation between connected nodes with different labels will bring noise for target node.

\begin{figure}
\centering
\includegraphics[width=0.5\textwidth]{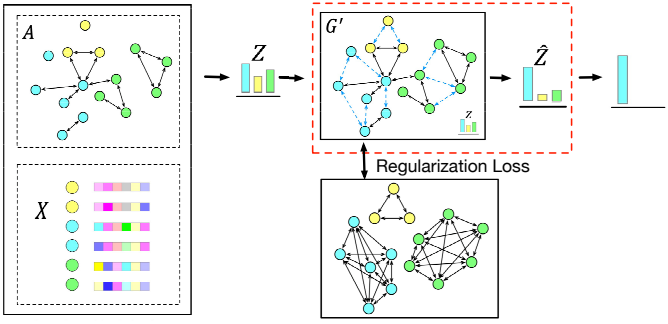}
\caption{Architecture of LC-GNN. Basic GNN model, e.g., GCN or GAT is first employed to learn a label distribution. Based on label distribution, an aggregation matrix is constructed and regularization loss is added to constrain the aggregation matrix.}
\label{fig:architecture}
\end{figure}

For graph-based semi-supervised learning, e.g., node classification, the extent to which connected nodes have the same label determines the performance of traditional graph neural networks. Neighboring nodes with different labels are harmful for node classification. For convenience, we define \emph{label consistency} as the proportion of neighboring nodes that have the same label with target node. Figure.~\ref{fig:realideal} shows the correlation between nodes's label consistency and classification accuracy when applying GCN to node classification on Cora dataset. The color of each node represents its label, and we can see that the classification accuracy increases with label consistency. This motivates us to leverage label consistency to improve traditional graph neural networks.

In this paper, we propose label-consistency based graph neural networks (LC-GNN), where the aggregation matrix is based on label-consistency. Instead of only relying on the graph structure, we first learn a label distribution for each node through traditional GNNs, and then calculate the aggregation weight between two nodes by measuring the similarity of their label distributions. These soft connections beyond the graph structure contribute to aggregating useful information from unconnected nodes with the same labels, weakening noise from connected nodes with different labels. Extensive experiments show that our model achieves state-of-the-art results in node classification, especially in sparse scenarios with only a handful of labeled nodes.

\section{Methods}

In this section, we propose label-consistency based graph neural networks, namely, LC-GNN. We build our new aggregation matrix based on learned label distribution, and aggregate information via this new aggregation matrix. Specifically, we first employ GCN or GAT to learn a label distribution for each node, capturing the feature information and local structure of each node. Based on these label distributions, we construct the new aggregation matrix and conduct label aggregation. Meanwhile, the label-consistency based aggregation matrix of nodes in the training set is leveraged as the regularization loss to constrain the learned aggregation matrix.

\subsection{Aggregation matrix based on label consistency}

Before constructing the aggregation matrix, we first employ GCN or GAT as the base models to learn a label distribution for each node, leveraging their capability to integrate node features and graph structure. As illustrated in Fig.~\ref{fig:architecture}, both the two models leverage $A$ and $X$, where  $A\in \mathbb{R}^{n\times n}$ be the adjacency matrix  and $X\in \mathbb{R}^{n\times p}$ denote the feature matrix. Then the basic models output the probability distributions $Z$ over labels for all the nodes. Formally, the probability distributions are calculated as follows.
\begin{equation}
Z=f(X,A)={\rm softmax}(\hat{A}{\rm ReLU}(\hat{A}XW^{(0)})W^{(1)}),
\label{eq:gcn}
\end{equation}
where $W^{(0)}\in \mathbb{R}^{p\times d}$ and $W^{(1)}\in \mathbb{R}^{d\times m}$ are weight matrices with $d$ as the dimension of the hidden layer. The first layer leverages rectified linear unit ($\rm ReLU$), and the second layer leverages $\rm softmax$ to obtain a probability distribution $Z\in \mathbb{R}^{n\times m}$. $Z$ is a row-normalized matrix, where each row represents the probability that the node belongs to corresponding labels.

We now define a novel aggregation matrix, leveraging label consistency among nodes. With the obtained label distribution $Z$, where $Z_{i,l}$ denotes the probability that node $i$ has label $l$, we calculate the label consistency between two nodes $i$ and $j$ as $ZZ^T$, the dot product of their label distributions. For the convenience of computation, we conduct row-normalization on $ZZ^T$ and obtain the label-consistency based aggregation matrix as
\begin{equation}
P=\mathrm{Row}\textit{-}\mathrm{Normalize}(Z{Z}^\top),
\label{eq:gp}
\end{equation}
where $\mathrm{Row}\textit{-}\mathrm{Normalize}(T)=Q^{-1}T$, and $Q\in \mathbb{R}^{n\times n}$ is a diagonal matrix with $Q_{ii}=\sum_{j}T_{ij}$. $P\in R^{n\times n}$ reflects the similarity of nodes based on label distribution. We regard it as our aggregation matrix.  This aggregation matrix based on label-distribution lays the solid foundation for our improvement over current models.

\subsection{Label-consistency based graph neural networks}

With the label-consistency based aggregation matrix $P$, label aggregation is conducted via 
\begin{equation}
\hat Z=PZ.
\label{eq:ld}
\end{equation}
 
 Note that the aggregation matrix $P$ actually defines a new graph $G'$ with $P$ as adjacency matrix, as shown in Fig. 2. Label aggregation corresponds to a kind of label propagation on $G'$.
 Different from previous methods, the new aggregation matrix can bridge any two nodes with similar label distributions. Consequently, $P$ enlarges the receptive field of each node and gets more nodes involved in the training stage. Specifically, only the first-order and second-order neighbors of training nodes are used to optimize a two-layer GCN or GAT. In contrast, all nodes can participate in training a two-layer LC-GCN or LC-GAT owing to the new connections beyond the local graph structure, and noise from some connected nodes with different labels  will be weakened.

\textbf{Loss Function}

The proposed LC-GNN is designed to solve semi-supervised node classification. The loss function consists of two parts, including cross-entropy over all labeled nodes and the regularization loss to constrain the learned aggregation matrix.
\begin{equation}
\mathcal{L} = \mathcal{L_C} + \lambda \mathcal{L_R},
\end{equation}
where $\lambda$ is the hyper-parameter to tune the weight of regularization loss.
The first part is cross-entropy loss function over all labeled nodes:
\begin{equation}
\mathcal{L_C}=-\sum_{i\in V^L} \sum_{l=1}^m Y_{il}\ln \hat Z_{il},
\end{equation}
where $V^L$ is the set of labeled nodes and $m$ is the number of labels, $Y_{il}$ is 1 if node $v_i$ has label $l$, otherwise 0.

We use label consistency among labeled nodes as regularization loss of the learned label consistency matrix. Denoting with $M$ the label consistency matrix among labeled nodes, we have $M_{ij}=1$ when nodes $i$ and $j$ have the same label, or otherwise $M_{ij}=0$. Since $M$ is un-normalized, we use it to regularize the un-normalized learned label consistency matrix $ZZ^\top$. Let $N$ denote $ZZ^\top$, the regularization loss is defined as cross-entropy over all pairs of labeled nodes:

\begin{equation}
\mathcal{L_R}=-\sum_{{i}\in V^L, {j}\in V^L} M_{ij} \ln N_{ij} +(1-M_{ij})\ln(1-N_{ij}).
\end{equation}

\subsection{Complexity Reduction}

Different from previous aggregation matrix, $P\in R^{n\times n}$ in Eq.~\ref{eq:gp} based on label-consistency is a dense matrix, thus storing $P$ and calculating Eq.~\ref{eq:ld} cost expensively. To combat this problem, we avoid explicit calculation of $P$ and thus reduces complexity.

\textbf{Theorem1:} given that $Z$ is a row-normalized matrix, then 
\begin{equation}
\mathrm{Row}\textit{-}\mathrm{Normalize}(Z{Z}^\top)Z = \mathrm{Row}\textit{-}\mathrm{Normalize}(Z ({Z}^\top Z)).
\label{eq:cal}
\end{equation}

\textbf{Proof:} The operator ``Row-Normalize'' can be replaced as multiplying with a diagonal-matrix multiplication, i.e., 
\begin{equation}
\mathrm{Row}\textit{-}\mathrm{Normalize}(Z{Z}^\top)Z=D^{-1}Z{Z}^\top Z,
\label{eq:rn1}
\end{equation}
\begin{equation}
\mathrm{Row}\textit{-}\mathrm{Normalize}(Z ({Z}^\top Z))=\hat D^{-1}Z{Z}^\top Z,
\label{eq:rn2}
\end{equation}
where $D$ and $\hat D$  are two diagonal matrices with $D_{i,i}=\sum_{j=1}^n [Z{Z}^\top]_{i,j}$ and $\hat D_{i,i}=\sum_{j=1}^n [Z{Z}^\top Z]_{i,j}$.  Let $T$ denote $Z{Z}^\top$, then we have
\begin{equation}
D_{i,i}=\sum_{j=1}^nT_{i,j}
\label{eq:rn3}
\end{equation}
\begin{equation}
\hat D_{i,i}=\sum_{k=1}^m\sum_{j=1}^nT_{i,j}Z_{j,k}=\sum_{j=1}^nT_{i,j}\sum_{k=1}^m Z_{j,k}=\sum_{j=1}^nT_{i,j}.
\label{eq:rn4}
\end{equation}
Because of the ``row-normalized'' property of $Z$, the diagonal matrix $D$ is the same as $\hat D$. As a result, the theorem is satisfied. Based on Theorem 1, we implement Eq.~\ref{eq:gp} and Eq.~\ref{eq:ld} via RHS of Eq.~\ref{eq:cal}, i.e., calculate ${Z}^\top Z\in R^{m \times m}$ firstly \textbf{to avoid the calculation of  $P$ with the size  $O(n^2)$}.

\section{Experiments}
To validate the proposed label-consistency based graph neural networks, we conduct experiments on three widely-used datasets for node classification task.

\subsection{Datasets}

The three benchmark datasets are Cora, Citeseer and PubMed~\cite{sen2008collective}. In these citation network datasets, nodes represent documents and edges are citation links. Table~\ref{table:Datasetstatistics} shows an overview of three datasets. Label rate denotes the proportion of labeled nodes for training.

\begin{table}[htbp]
\centering
\caption{Statistics of Datasets}
\small
\begin{tabular*}{\hsize}{@{}@{\extracolsep{\fill}}l r r r r r r@{}}
\hline
\textbf{Datasets} & \textbf{Nodes} & \textbf{Edges} & \textbf{Classes} & \textbf{Features} & \textbf{Label Rate}\\
\hline 
Cora & 2,708 & 5,429 & 7 & 1,433 & 0.052\\
Citeseer  & 3,327 & 4,732 & 6 & 3,703 & 0.036\\
PubMed & 19,717 & 44,338 & 3 & 500 & 0.003\\
\hline
\end{tabular*}
\label{table:Datasetstatistics} 
\end{table}

\subsection{Baselines}
We compare with traditional graph semi-supervised learning methods, including feature-based Multi-Layer Perceptron (MLP), label propagation (LP)~\cite{zhu2003semi}, semi-supervised embedding (SemiEmb)~\cite{weston2012deep}, manifold regularization (ManiReg)~\cite{belkin2006manifold}, graph embeddings (DeepWalk)~\cite{perozzi2014deepwalk}, iterative classification algorithm (ICA)~\cite{lu2003link} and Planetoid~\cite{yang2016revisiting}.
Furthermore, since graph neural networks are proved to be effective in semi-supervised learning on graphs, we also compare against the representative graph neural networks, i.e., ChebyNet~\cite{defferrard2016convolutional}, GCN~\cite{kipf2017semi}, MoNet~\cite{monti2017geometric} and GAT~\cite{velickovic2017graph}.

We implement our LC-GNN model using GCN and GAT as base models, and the resulted two models are referred to as LC-GCN and LC-GAT respectively. To offer a fair comparison with base models, we also implement them using the same setting as LC-GCN and LC-GAT, obtaining GCN* and GAT*. Finally, to validate the effect of the two components in our models, i.e., label consistency (LC) and regularization loss (RL), we also implement variants of our models and conduct ablation analysis, i.e., without both LC and RL (w/o LC, w/o RL), or only without RL (w/o RL). Note that there is no variant only without LC, since RL depends on LC. For without LC, we use the adjacency matrix to implement label aggregation.

\subsection{Experimental Settings}
We implement our models using the PyTorch-Geometric library \cite{fey2019fast}, and follow the settings in \cite{fey2019fast} to pre-train GCN and GAT. 
Next, we train LC-GCN and LC-GAT and do label aggregation based on label-consistency and feature aggregation based on local structure as aforementioned. The partition of datasets is the same as
GCN~\cite{kipf2017semi} with an additional validation set of 500 labeled samples to determine hyper-parameters. We use Adam optimizer with an initial learning rate of 0.01 and a weight decay of 0.0005. For LC-GAT, we set the learning rate as 0.005.
The hyper-parameter $\lambda$ is set to be $2.0$ in Cora and $1.0$ in Citeseer and PubMed.
We run 1000 epochs and choose the model that performs the best on the validation set. 
We report the classification accuracy on the test set.

\subsection{Performance on Node Classification Task}

\begin{table}[!t]
\small
\centering
\caption{Results of Node Classification (Fixed Partition)}
\begin{tabular*}{\hsize}{@{}@{\extracolsep{\fill}}l l l l@{}}
\hline
\textbf{Method} & \textbf{Cora} & \textbf{Citeseer} & \textbf{PubMed} \\
\hline
MLP & 55.1\% & 46.5\% & 71.4\% \\
LP & 68.0\% & 45.3\%& 63.0\% \\
SemiEmb & 59.0\% & 59.6\%& 71.7\% \\
ManiReg & 59.5\% &60.1\% & 70.7\% \\
DeepWalk & 67.2\% & 43.2\%& 65.3\% \\
ICA & 75.1\% & 69.1\%& 73.9\% \\
Planetoid & 75.7\% & 64.7\%& 77.2\% \\
\hline
ChebyNet & 81.2\% & 69.8\%& 74.4\% \\
GCN & 81.5\% & 70.3\%& 79.0\% \\
MoNet & 81.7$\pm$0.5\% & --- & 78.8$\pm$0.3\% \\
GAT & 83.0$\pm$0.7\% & 72.5$\pm$0.7\% & 79.0$\pm$0.3\% \\
\hline
GCN* & 81.2$\pm$0.6\% & 71.1$\pm$0.5\% & 78.9$\pm$0.6\%\\
LC-GCN (w/o LC, w/o RL) & 81.1$\pm$0.4\% & 70.3$\pm$0.6\% &  79.0$\pm$0.5\%\\
LC-GCN(w/o RL) & 82.5$\pm$0.4\% & \textbf{72.3}$\pm$0.9\% & 79.9$\pm$0.4\%\\
LC-GCN & \textbf{82.9}$\pm$0.4\% & \textbf{72.3}$\pm$0.8\% &  \textbf{80.1}$\pm$0.4\%\\
\hline
GAT*  & 83.2$\pm$0.4\% & 71.1$\pm$0.7\% & 78.9$\pm$0.4\%\\
LC-GAT(w/o LC, w/o RL) & 83.0$\pm$0.5\% &70.6$\pm$0.6\% &  77.7 $\pm$0.5\%\\
LC-GAT(w/o RL) & 83.0$\pm$0.6\% & \textbf{73.8}$\pm$0.5\% &  76.0$\pm$0.5\%\\
LC-GAT  & \textbf{83.5}$\pm$0.4\% & \textbf{73.8}$\pm$0.7\% & \textbf{79.1}$\pm$0.5\%\\
\hline
\end{tabular*}
\label{table:results-1}
\end{table}

We now validate the effectiveness of LC-GCN and LC-GAT on node classification. Similar to previous methods, we report the mean classification accuracy metric (with standard deviation) for quantitative evaluation on three citation networks. Experimental results are reported in Table~\ref{table:results-1}. Bold numbers indicate that our method improves the base model, i.e., GCN* and GAT*.

Graph neural network methods all perform much better than traditional methods, i.e., feature-based methods and network embedding methods. This is due to that graph neural networks are trained in an end-to-end manner, and update representations via graph structure under the guide of labels. LC-GCN(w/o RL) and LC-GAT(w/o RL) achieve an improvement over GCN and GAT due to the label consistency based aggregation matrix.  The result of LC-GAT(w/o RL) on PubMed drops a little, which may result from feature sparseness on PubMed. Furthermore, LC-GCN and LC-GAT achieve an improvement over LC-GCN(w/o RL) and LC-GAT(w/o RL), showing the effectiveness of regularization loss.

\label{sec:exp}
We now analyze why LC-GCN and LC-GAT outperform over their base models. For this purpose, we show the performance of these methods on Cora (Table~\ref{table:results-few}) and Citeseer (Table~\ref{table:results-fewc}), varying the number of labeled nodes. Experimental results demonstrate that the superiority of LC-GCN and LC-GAT over their base models increases when the number of labeled nodes decreases. This indicates that our proposed methods are promising in semi-supervised node classification, especially when labeled examples are time-consuming or difficult to obtain. The advantage of our models takes roots in leveraging label-consistency to enlarge the receptive fields of nodes for information aggregation.

\begin{table}[h]
\caption{Performance in sparse scenarios on Cora}
\begin{tabular*}{\hsize}{@{}@{\extracolsep{\fill}}l l l l@{}}
\hline
\textbf{Method} & \textbf{5 labels} & \textbf{10 labels} & \textbf{15 labels} \\
\hline
GCN & 69.4$\pm$2.8\% & 73.8$\pm$1.0\% &  79.6$\pm$0.6\%\\
LC-GCN & \textbf{76.3}$\pm$1.6\% & \textbf{77.7}$\pm$1.0\% &  \textbf{82.1}$\pm$0.8\%\\
GAT  & 77.0$\pm$0.4\% & 77.3$\pm$0.6\% & 81.8$\pm$0.5\%\\
LC-GAT  & \textbf{77.7}$\pm$1.0\% & \textbf{78.7}$\pm$0.6\% & \textbf{82.4}$\pm$0.3\%\\
\hline
\end{tabular*}
\label{table:results-few}
\end{table}

\begin{table}[h]
\caption{Performance in sparse scenarios on Citeseer}
\begin{tabular*}{\hsize}{@{}@{\extracolsep{\fill}}l l l l@{}}
\hline
\textbf{Method} & \textbf{5 labels} & \textbf{10 labels} & \textbf{15 labels} \\
\hline
GCN & 52.3$\pm$1.9\% & 66.4$\pm$1.2\% &  68.6$\pm$0.7\%\\
LC-GCN & \textbf{69.1}$\pm$1.2\% & \textbf{70.3}$\pm$1.0\% &  \textbf{71.0}$\pm$0.4\%\\
GAT  & 58.0$\pm$1.2\% & 67.6$\pm$0.8\% &69.2$\pm$0.5\%\\
LC-GAT  & \textbf{69.0}$\pm$0.5\% & \textbf{70.8}$\pm$0.3\% & \textbf{71.4}$\pm$0.7\%\\
\hline
\end{tabular*}
\label{table:results-fewc}
\end{table}

\section{Conclusion}
Previous methods follow a ``neighborhood aggregation'' mechanism based on the local structure to aggregate information. In this paper, we propose label-distribution based graph neural networks. We first build our new aggregation matrix based on learned label distribution. Then we aggregate information via this new aggregation matrix. The mechanism is applicable to current models. Extensive experiments and analysis demonstrate our model achieves best results in the task of graph-based semi-supervised node classification.

\section{Acknowledgments}
This paper is funded by the National Natural Science Foundation of China under Grant Nos. 91746301 and 61425016. Huawei Shen is also funded by K.C. Wong Education Foundation and Beijing Academy of Artificial Intelligence (BAAI). 

\bibliographystyle{abbrv}
\bibliography{sample-base}

\end{document}